\RequirePackage{fix-cm}
\documentclass[smallextended]{svjour3}       
\smartqed  
\usepackage{graphicx}

\usepackage{subfig}

\usepackage{framed,multirow}

\usepackage{amssymb}
\usepackage{latexsym}
\usepackage{subfig}
\begin{document}

\title{Ensemble emotion recognizing with multiple modal physiological signals
}
\author{Jing Zhang\and
        Yong Zhang\footnote{Corresponding Author} \and
        Suhua Zhang\and
        Cheng Cheng
}


\institute{Jing Zhang \at
              School of computer and information technology, Liaoning Normal University, Liushu south street NO.1 Ganjingzi district, Dalian and 116018, China \\
              Tel.: +86-0411-85992418\\
              Fax: +86-0411-85992418\\
              \email{zhangjing\_0412@163.com}
           \and
             Yong Zhang\at
              School of computer and information technology, Liaoning Normal University, Liushu south street NO.1 Ganjingzi district, Dalian and 116018, China\\
              \email{ayong\_zh@163.com}
}

\date{Received: date / Accepted: date}

\maketitle

\begin{abstract}
Physiological signals that provide the objective repression of human affective states are attracted increasing attention in the emotion recognition field. However, the single signal is difficult to obtain completely and accurately description for emotion. Multiple physiological signals fusing models, building the uniform classification model by means of consistent and complementary information from different emotions to improve recognition performance. Original fusing models usually choose the particular classification method to recognition, which is ignoring different distribution of multiple signals. Aiming above problems, in this work, we propose an emotion classification model through multiple modal physiological signals for different emotions. Features are extracted from EEG, EMG, EOG signals for characterizing emotional state on valence and arousal levels. For characterization, four bands filtering theta, beta, alpha, gamma for signal preprocessing are adopted and three Hjorth parameters are computing as features. To improve classification performance, an ensemble classifier is built. Experiments are conducted on the benchmark DEAP datasets. For the two-class task, the best result on arousal is 94.42\%, the best result on valence is 94.02\%, respectively. For the four-class task, the highest average classification accuracy is 90.74, and it shows good stability. The influence of different peripheral physiological signals for results is also analyzed in this paper.
\keywords{Emotion recognition \and physiological signal processing \and ensemble classifier }
\end{abstract}

\section{Introduction}
\label{intro}
Emotion is a very important part to each of us, affecting our daily life such as work efficiency, learning status, social contact and decision making. Recognizing emotions through speech [1], facial expressions [2] and gestures [3] are very significant components of research works in this field. However, all of these can be artificially changed or controlled, sometimes the results from them don¡¯t really reflect person¡¯s emotion status. Hence, more and more researchers pay attention to conduct emotion recognition by processing physiological signals, which are more objective.
Due to the recent advancement in building wireless and wearable sensors [4], physiological signals are relatively easy to acquire without damage to body. Human body is a combination of various organs, for example heart, brain, muscles and eyes. Some organs produce electrical signals, such as the electroencephalogram (EEG) produced by the brain, electrooculography (EOG) produced by eyes, electromyogram (EMG) produced by muscles, and blood volume pressure (BVP). Most of these signals have been applied to many fields, such as building a wireless monitoring system to help patients and elders to monitor their own physiological conditions in real time [5], to detect the stress state of the subjects [6], to monitor the sleep quality [7], and so on. Among these signals, EEG has received considerable attentions from researchers in the field of emotion recognition. Alarcao and Fonseca [8] provided a comprehensive overview of the work published between 2009 and 2016 on emotion recognition using EEG signals, and gave some suggestions to achieve reproducible, replicable, well-validated and high-quality results. When classifying EEG signals, existing studies have indicated that peripheral physiological signals can also change with emotion [9]. Jerritta et al. [10] extracted high-order statistical (HOS) features from facial EMG signals and mapped these features into corresponding emotions using k-nearest neighbor classifier. Nabian et al. [11] presented the development of a biosignal-specific processing toolbox (Bio-SP tool) for preprocessing and feature extraction of physiological signals, such as EOG, EMG, electrocardiogram (ECG), electrodermal activity, continuous blood pressure.
However, some studies have found that it is very difficult to accurately reflect emotion status using a single physiological signal [12], and more and more studies attempt to achieve better classification performance by co-processing multiple features [13-15]. Wen et al. [16] used galvanic skin response (GSR), fingertip oxygen saturation and heart rate as input signals to classify five emotions through random forests. Das et al. [17] combined ECG and GSR signals and calculated their power spectral density (PSD) as features for the classification of three emotions: happy, sad, and neutral.
The emotion classification problems consist of processes such as preprocessing, feature extraction, classification and analysis [18]. In the survey [8], approximately 84\% of the works they collected used some band pass filters as the preprocessing method, reflecting the importance of different bands, such as delta, theta, alpha, beta and gamma. The feature extraction process can be handled using various methods, among which the Fourier transform, PSD and entropy are widely used [8]. Fourier transform includes the short-time Fourier transform (STFT) and discrete Fourier transform (DFT). Entropy includes approximate entropy (AE), sample entropy (SE), differential entropy (DE), and wavelet entropy (WE). In addition, there are other methods also applied to extract features, like, wavelet transform [19], empirical mode decomposition (EMD) [20], auto-regressive (AR) [21], and so on.
In addition to data preprocessing and feature extraction, classification phase is an important part of emotion recognition model. There are plenty of classifiers for automatic emotion identification [22], such as support vector machine (SVM), K-nearest neighbor (KNN), linear discriminant analysis (LDA), and naive Bayes [8]. Most of the above methods use a single classifier to recognize emotions. Although a single classifier can achieve good recognition results, recent research shows that deep recognition models [23-26], and combination of multiple classifiers, i.e., ensemble learning, can get better results. Many ensemble strategies have been proposed [27], such as bagging, boosting, and stacking.
In this paper, we propose an emotion classification ensemble model for emotion classification problem. Our analysis mainly focused on the combination of different peripheral physiological signals with EEG, and the impact of integration of multiple classifiers on results. The performance of proposed method is investigated on DEAP emotion database [28]. The experimental details and results will be shown and compared in the experimental sections.
The rest of paper is organized as follows. Section 2 briefly introduces an emotion analysis database used in this paper. In Section 3, an emotion recognition model through multimodal physiological signals for different emotions is proposed. And the experimental results and analysis are given in Section 4. Section 5 concludes this paper.

\section{Multiple modal physiological signals}

\subsection{Physiological signals description}
\label{sec:2}
The pre-processed data set from database for emotion analysis using physiological signals DEAP [28] is used in our research. The database contains 32 subjects¡¯ physiological signals which were got from 40 channels, 32 channel EEG data were recorded using a Biosemi ActiveTwo system and 8 channel peripheral physiological signals were recorded around the body using sensors, including hEOG, vEOG, zEMG, tEMG, GSR, respiration belt, plethysmograph and temperature. Afterwards, the data was down-sampled to 128Hz from 512Hz, and eye blink artifact removal via independent component analysis. During collection, each subject was presented with forty, one-minute long music videos with varying emotional content. Then she/he was asked to fill a self-assessment for her/his valence, arousal, liking and dominance from 1 to 9.
A standard for evaluating and comparing accuracies of emotion recognition methods is not established now. The selection of EEG electrode channels and time segments are always a controversial problem. In this research, to avoid the loss of information, we used the whole data from EEG channels. For peripheral physiological signals, hEOG and vEOG are merged called EOG, zEMG and tEMG are merged called EMG. They are combined with EEG data to classify emotion state. The accuracies from them separately are compared and discussion in section IV.
\subsection{Emotion model}
Psychologists tend to divide emotion models into two categories, discrete emotion models and multi-dimensional emotion space model. For the multi-dimensional emotion model is more persuasive in explaining the degree of people's emotional differences and the process of mutual transformation between emotions, it has been used by more and more researchers. DEAP takes a multi-dimensional emotion space model, including valence, arousal, dominance and liking axes. The affective states are measured using two dimensions (valence, arousal) in our paper. The valence ranges from unpleasant to pleasant, while arousal ranges from passive to active. The valence-arousal scale model explains emotion variation in a 2D plane, which is divided into four regions: Low Valence-Low Arousal (LVLH), Low Valence-High Arousal (LVHA), High Valence-Low Arousal (HVLA) and High Valence-High Arousal (HAHV). Emotional state definition is converted into determining valence and arousal levels. Our research is carried out for the two-class task and the four-class task separately. TABLE 1 and TABLE 2 shows the number of samples for each class in DEAP database.
\begin{table}
\caption {\leftline{Total 1280 samples for two-class task}}
\label{Total 1280 samples for two-class task}       
\begin{tabular}{lll}
\hline\noalign{\smallskip}
Actual Labels & Arousal & Valence  \\
\noalign{\smallskip}\hline\noalign{\smallskip}
Low & 526 & 556 \\
High & 754 & 724 \\
\noalign{\smallskip}\hline
\end{tabular}
\end{table}

\begin{table}
\caption{\leftline{Total 1280 samples for four-class task}}
\label{tab:2}       
\begin{tabular}{lll}
\hline\noalign{\smallskip}
Actual Labels & Low Valence & High Valence \\
\noalign{\smallskip}\hline\noalign{\smallskip}
Low Arousal & 260 & 266 \\
High Arousal & 296 & 458 \\
\noalign{\smallskip}\hline
\end{tabular}
\end{table}
\section{The proposed model}
This section mainly focuses on the method of our emotional recognition model for each phase of preprocessing, feature extraction and classification. The specific procedure of our proposed method is showed in Fig. 1.
\subsection{Pre-processing}
Considering different signal bands contain characteristic of emotion, it is very necessary to filter different bands from physiological signals to extract more targeted features for improving the final classification performance. The frequency of signals contained in DEAP is from 4Hz to 45Hz. In our paper, theta (4-8Hz), alpha(8-13Hz), beta(13-30Hz) and gamma(30-43Hz) bands are filtered from different physiological signals respectively by adopting Butterworth filters [29].
%
\begin{figure*}
  \includegraphics[width=0.75\textwidth]{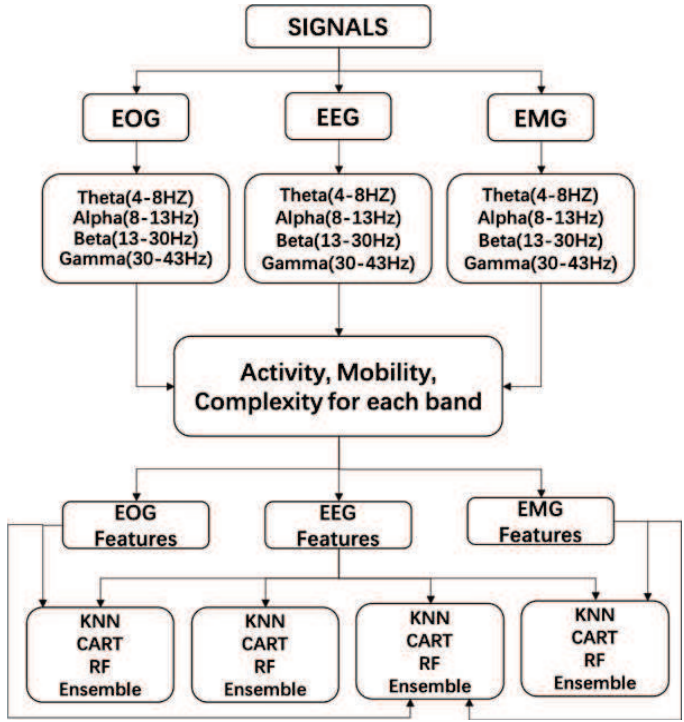}
\caption{\leftline{The flow diagram of proposed method}}
\label{fig:1}       
\end{figure*}
Butterworth filters have a magnitude response that is maximally flat in the pass band and monotonic overall. This smoothness comes at the price of decreased roll off steepness. Its low pass filter squared amplitude response can be represented as equation (1).
\begin{equation}
{\left\| {{H_a}\left( {j\omega } \right)} \right\|^2} = \frac{1}{{1 + {{(\frac{\omega }{{{\omega _c}}})}^{2N}}}}
\end{equation}
where $N$ is the order of filter. In our experiment, the value is set to 8. ${\omega _c}$ is the cut-off frequency at which $|{H_a}\left( {j\omega } \right)| = 1/\sqrt 2$. In general, ${\omega _c}$ is also known as the -3dB cut-off frequency.
\\ To simulate band pass filter to get the specific band mentioned above, we first consider the transfer function ${H_a}\left( s \right)$ in the s-domain as shown in equation (2), which is the same as ${H_a}\left( {j\omega } \right)$ when $s = j\omega $.
\begin{equation}
{H_a}(s) = \frac{{\omega _c^N}}{{\mathop \prod \nolimits_{k = 0}^{N - 1} (s - {s_k})}}
\end{equation}
where ${s_k}\;\left( {k = 0,1, \ldots ,N - 1} \right)$ is a pole in the s-plane.
\\ Since the amplitude-frequency characteristics of the filters are different, all frequencies are normalized in order to unify the design. Let $\lambda  = \frac{\omega }{{{\omega _c}}}$, where $\lambda$  is normalized frequency, and $p = j,\lambda  = \frac{{j\omega }}{{{\omega _c}}}$, where $p$ is normalized complex variable. Now the normalized Butterworth transfer function can be written as equation (3).
\begin{equation}
{H_a}\left( p \right) = \frac{1}{{\mathop \prod \nolimits_{k = 0}^{N - 1} \left( {p - {p_k}} \right)}}
\end{equation}
where ${p_k}\;\left( {k = 0,1, \ldots ,N - 1} \right)$ is a normalized pole.
\\ To change low pass filter to band pass filter, the index conversion is done. Let ${\omega _u}$ be pass band upper limit frequency, ${\omega _l}$ be pass band lower limit frequency, ${\omega _0}$ be pass band center frequency. And let: $B = {\omega _u} - {\omega _l}$, ${\eta _l} = \frac{{{\omega _l}}}{B}$, ${\eta _u} = \frac{{{\omega _u}}}{B}$, $\eta _0^2 = {\eta _l}{\eta _u}$. So $\lambda  = \frac{{{\eta ^2} - \eta _0^2}}{\eta }$.
\\ Through the index conversion we can get the band pass filter transfer function as shown in equation (4). By passing the specific four bands parameters to it, we get each of the four bands to extract features.
\begin{equation}
{\left. {{H_a}\left( s \right) = {H_a}\left( p \right)} \right|_{p = \frac{{{s^2} + {\omega _l}{\omega _u}}}{{s\left( {{\omega _u} - {\omega _l}} \right)}}}}
\end{equation}
\subsection{Features extraction}
Hjorth proposed the Hjorth parameter in 1970 [30], which provides a fast method for calculating three important features of signals in time domain, including Activity, Mobility, and Complexity. It has been widely used in physiological signal processing area. After filtering out of different brain rhythms, features are extracted by calculating the parameters from each of them.
\\ The activity parameter represents the signal power, which can indicate the surface of power spectrum in the frequency domain. It can be calculated by equation (5).
\begin{equation}
Activity = {\mathop{\rm var}} (y(t))
\end{equation}
where $y(t)$ represents the signal.
\\ The mobility parameter represents the mean frequency or the proportion of standard deviation of the power spectrum. This is defined as the square root of variance of the first derivative of the signal $y(t)$ divided by variance of the signal $y(t)$. It is denoted as equation (6).
\begin{equation}
Mobility = \sqrt {\frac{{{\mathop{\rm var}} (\frac{{dy(t)}}{{dt}})}}{{{\mathop{\rm var}} (y(t))}}}
\end{equation}
\\ The Complexity parameter represents the change in frequency. The parameter compares the signal's similarity to a pure sine wave, where the value converges to 1 if the signal is more similar. It is represented by the following equation (7).
\begin{equation}
Complexity = \sqrt {\frac{{Mobility\left( {\frac{{dy(t)}}{{dt}}} \right)}}{{Mobility\left( {y(t)} \right)}}}
\end{equation}
\\ The Hjorth parameters based on variance have faster calculation speed than other methods. We calculate the three parameters from EEG, EOG, EMG, and combine them to form feature sets for classification.
\subsection{Ensemble classification}
Most of the emotion recognition model takes only one classifier to get the final results. It does not have enough stability and high accuracy. Bagging (Bootstrap Aggregating) is an ensemble meta-algorithm, which uses bootstrap sampling for training data, that is, sampling data is returned. Each time the sampling dataset trains a base classifier, and the number of sub-samples corresponds to the number of base classifiers obtained. The classification result of the base classifier is combined according to the principle of majority vote.
We establish the ensemble classifier by KNN, Rand Forest (RF) and CART based on bagging, one of parallel integrated classification methods. Although bagging is usually applied to decision tree methods, it can be used with any type of method. A single KNN classifier has got good results in other studies, more than 80 [31]. KNN cooperates with classic data mining method RF, CART to vote can improve the accuracy. The classification results of SVM are not balanced for the class tasks, accuracy on the fewer class is too low. The results for each single classifier and their ensemble classifier are displayed in the section IV.

\section{Experiments and analysis}
\subsection{Experiments setup}
The experiments are divided into two parts: two-class task and four-class task on the valence-arousal scale model.
To validate the effectiveness of our model and analyze the impact of peripheral physiological signals, four combinations of physiological signals are used, namely, EEG only, EEG with EOG, EEG with EMG, EEG with EOG and EMG.
\\There are 7680 (128*60) data points for each sample¡¯s per channel. EEG collected 32 channels, while both EOM and EMG collected two channels, respectively. After filtering each row data to four bands, namely theta, alpha, beta and gamma, 30720 (7680*4) data points are obtained for one channel. Obviously, it is difficult to classify directly due to massive data. To reduce the feature dimensions and get effective features, we compute Hjorth parameters in time domain after processing signals in frequency domain.
\\For each sample, each channel has 60 seconds data to deal with. During the calculation of the features, the fixed time window of each segment is 10 seconds. That is to say, every ten seconds of data can be calculated into three parameters. So, we can get 2304 features from each band of each channel for one sample to classify. EEG features and the combination of EEG with EOG features, EMG features are input into the ensemble classifier respectively. Besides ensemble classifier, each base classifier, KNN, RF, CART, is also experimented to compare with each other. The results are shown and analysized in the following section.
\\In the ensemble classifier, one test sample can get three test labels from three base classifiers. We adopt the method of majority vote to determine the class it ultimately belongs to.
\subsection{Experimental results and discussion}
\begin{description}
  \item(1) Performance evaluation parameters
\end{description}
The accuracy rate is one of the most commonly used evaluation parameters in classification. The accuracy pacc is defined as:
\[{p_{acc}} = ({n_{TN}} + {n_{TP}})/({n_{TN}} + {n_{FN}} + {n_{TP}} + {n_{FP}})\]
where ${n_{TN}}$ is the number of correctly predicted high-level instances, $n_{TP}$ is the number of correctly predicted low-level instances, $n_{FN}$ and $n_{FP}$ are the number of wrongly classified instances for high-level and low-level respectively. Considering the class imbalance, for two-class task F-score is also calculated as an evaluation parameter:
\[F - score = 2{\rm{}}{P_{pre}}{\rm{}}\frac{{{P_{sen}}}}{{{P_{pre}} + {P_{sen}}}}\]
where ${P _pre}$ is defined as ${P_{pre}} = \frac{{{n_{TP}}}}{{{n_{TP}} + {n_{FP}}}}$, ${P _sen}$ is defined as ${P_{sen}} = \frac{{{n_{TP}}}}{{{n_{TP}} + {n_{FN}}}}$.
\begin{description}
  \item(2) Two-class task
\end{description}
The emotion recognition classification results for two-class task are displayed in Table 3. From Table 3, we can obtain the following observations: First, compared with single base classifier, the ensemble classifier performs best, and the corresponding result outperforms those mentioned above. Second, the addition of peripheral physiological signals can enhance the classification results of EEG signals, but not obviously.
\begin{table}
\caption{\leftline{Results of Experiments for Two-class Task}}
\label{tab:3}       
\begin{tabular}{lllll}
\hline\noalign{\smallskip}
\multirow{2}{*}{Methods} & \multicolumn{2}{c} {Low/High Arousal}  & \multicolumn{2}{c} {Low/High Valence}\\
 & P\_acc & F-score & P\_acc & F-score \\
\noalign{\smallskip}\hline\noalign{\smallskip}
EEG+KNN & 79.98\% $\pm$ 3.28\% & 0.7546 & 78.8\% $\pm$ 3.09\% & 0.7575\\
EEG+CART & 93.35\% $\pm$ 2.51\% & 0.9186 & 92.74\% $\pm$ 2.47\% & 0.9171\\
EEG+RF & 87.97\% $\pm$ 3.32\% & 0.8528 & 87.83\% $\pm$ 3.18\% & 0.8599\\
EEG+ENS & 94.04\% $\pm$ 2.29\% & 0.9265 & 93.80\% $\pm$ 2.38\% & 0.9284\\
EEG+EOG+KNN & 80.32\% $\pm$ 3.71\% & 0.7596 & 79.17\% $\pm$ 3.53\% & 0.7622\\
EEG+EOG+CART & 93.39\% $\pm$ 2.37\% & 0.9194 & 92.62\% $\pm$ 2.40\% & 0.9153\\
EEG+EOG+RF & 87.77\% $\pm$ 3.27\% & 0.8497 & 88.24\% $\pm$ 3.31\% & 0.8644\\
EEG+EOG+ENS & 94.33\% $\pm$ 2.26\% & 0.9299 & 94.20\% $\pm$ 2.17\% & 0.9329\\
EEG+EMG+KNN & 80.53\% $\pm$ 3.20\% & 0.7617 & 79.44\% $\pm$ 3.43\% & 0.7644\\
EEG+EMG+CART & 93.50\% $\pm$ 2.59\% & 0.9206 & 93.12\% $\pm$ 2.30\% & 0.9210\\
EEG+EMG+RF & 87.66\% $\pm$ 3.60\% & 0.8492 & 88.12\% $\pm$ 3.05\% & 0.8625\\
EEG+EMG+ENS & 94.41\% $\pm$ 2.16\% & 0.9308 & 93.69\% $\pm$ 2.26\% & 0.9272\\
EEG+EOG+EMG+KNN & 80.52\% $\pm$ 3.16\% & 0.7615 & 79.10\% $\pm$ 3.42\% & 0.7609\\
EEG+EOG+EMG+CART & 93.53\% $\pm$ 2.64\% & 0.9210 & 93.06\% $\pm$ 2.35\% & 0.9204\\
EEG+EOG+EMG+RF & 87.49\% $\pm$ 3.67\% & 0.8475 & 87.77\% $\pm$ 3.48\% & 0.8596\\
EEG+EOG+EMG+ENS & 94.42\% $\pm$ 1.96\% & 0.9310 & 94.02\% $\pm$ 2.15\% & 0.9308\\
\noalign{\smallskip}\hline
\end{tabular}
\end{table}

\begin{figure*}
  \includegraphics[width=1.0\textwidth]{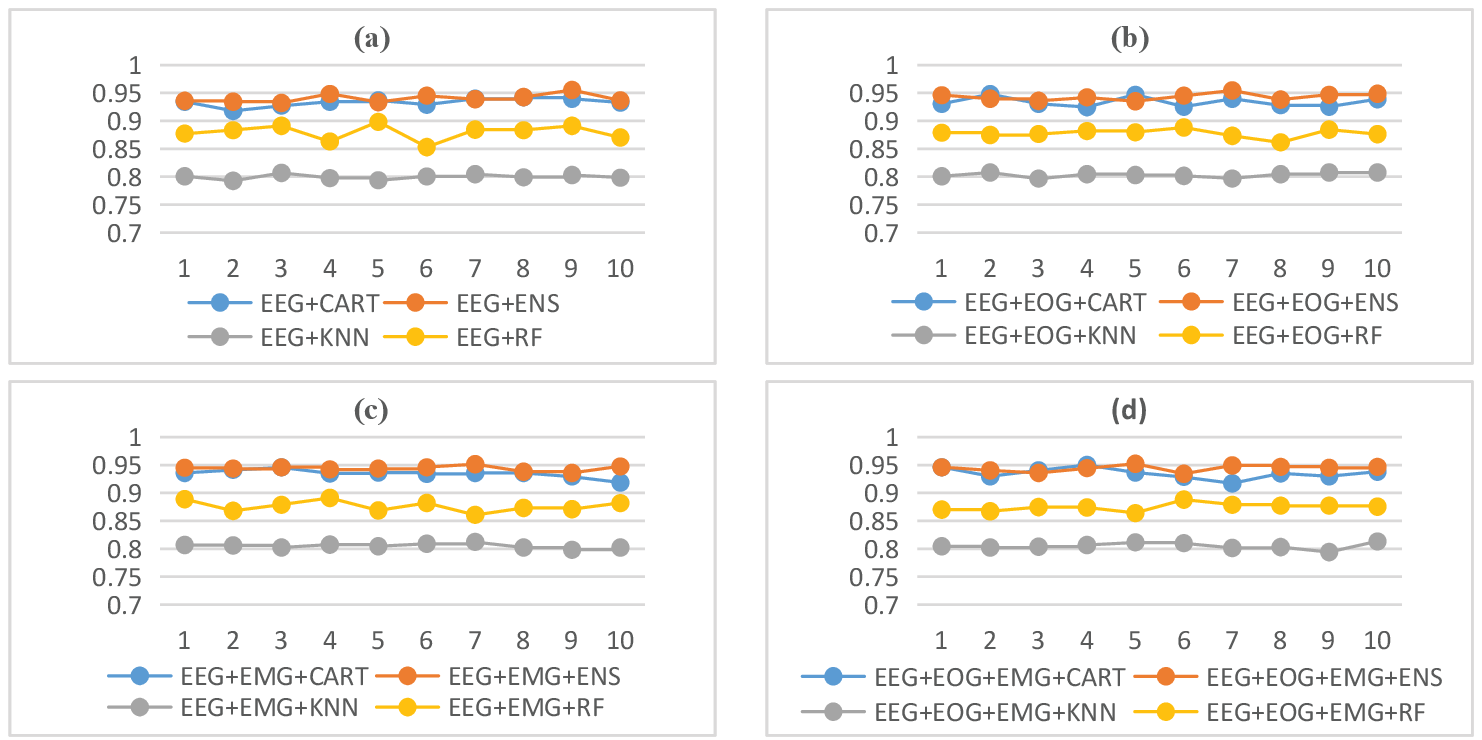}
\caption{The results of different classifiers on arousal dimension. (a) EEG signal on different classifiers; (b) the combination of EEG and EOG signals on different classifiers; (c) the combination of EEG and EMG signals on different classifiers; (d) the combination of EEG, EOG and EMG signals on different classifiers.}
\label{fig:2}       
\end{figure*}

\begin{figure*}
  \includegraphics[width=1.0\textwidth]{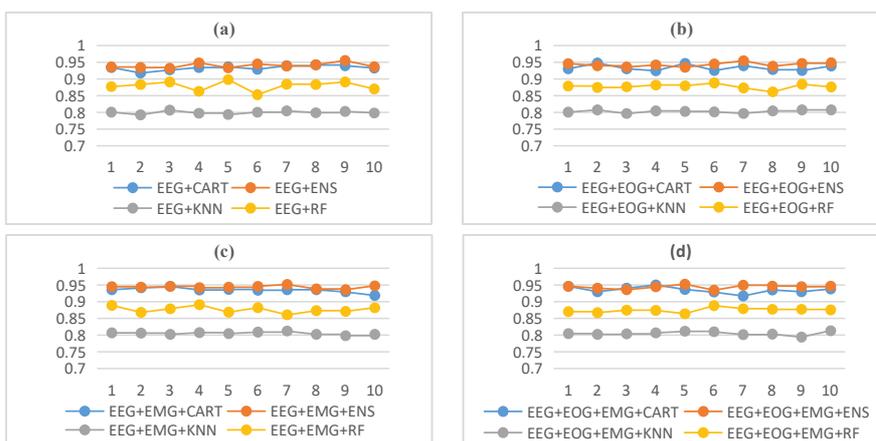}
\caption{The results of different classifiers on valence dimension. (a) EEG signal on different classifiers; (b) the combination of EEG and EOG signals on different classifiers; (c) the combination of EEG and EMG signals on different classifiers; (d) the combination of EEG, EOG and EMG signals on different classifiers.}
\label{fig:3}       
\end{figure*}
The focus of this research was to do emotion classification using EEG, EOG, EMG signals on arousal and valence dimension, and comparing results by different combinations with different classifier. The best accuracies are 94.42 and 94.02, obtained respectively by our ensemble classifier via EEG enhanced by EOG and EMG on arousal and valence dimensions. And this combination of signal and classifier performs most stable. From Fig.2 and Fig.3, we can find that the ensemble classifier has a very good classification accuracy for each class. CART performs good as well, whereas, by calculating standard deviation, the result is more fluctuating than ensemble classifier in Table 3. RF and KNN do not show good classification performance for this problem during our experiments.
\\ For different combination of physiological signals, EEG with EOG and EMG have best results, the join of peripheral physiological signals makes the result more stable than EEG only. Whereas, from Fig.4 and Fig.5, the influence of different classifiers on the classification effect is much greater than the combinations of the different signals. Table 4 shows some existing researches for two-class task. \\ Zoubi et al. [32] identified the human emotional state through the LSM model with the accuracy of 88.54 on arousal and 84.63 on valence. Piho and Tjahjadi [33] conducted emotion recognition by shortening the signal to find the strongest part of the mutual information and obtained the result of 89.84 on arousal and 89.61 on valence. Therefore, the methods proposed in this paper get higher results than existing methods. We can see that our model has the best performance.
\begin{figure*}
  \includegraphics[width=1.0\textwidth]{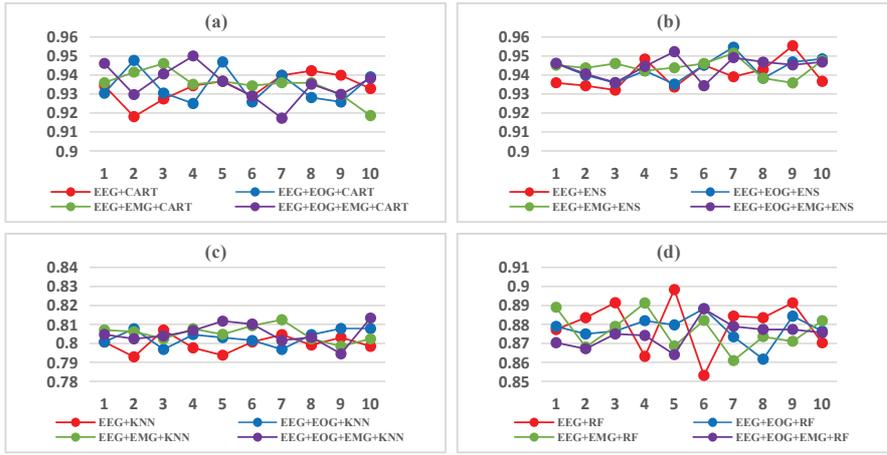}
\caption{The results of different physiological signals on classifiers for arousal dimension. (a) different physiological signals on CART; (b) different physiological signals on ENS; (c) different physiological signals on KNN; (d) different physiological signals on RF.}
\label{fig:4}       
\end{figure*}

\begin{figure*}
  \includegraphics[width=1.0\textwidth]{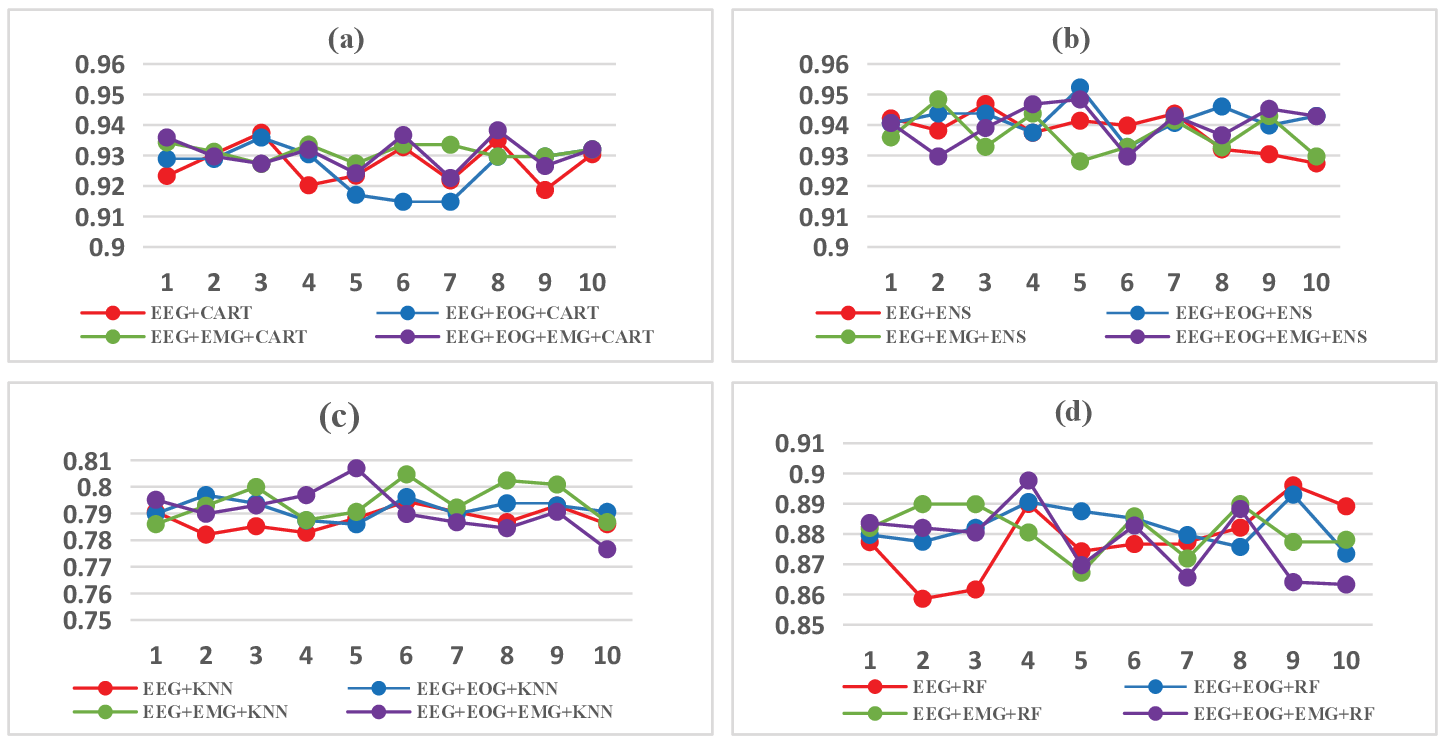}
\caption{The results of different physiological signals on classifiers for valence dimension. (a) different physiological signals on CART; (b) different physiological signals on ENS; (c) different physiological signals on KNN; (d) different physiological signals on RF.}
\label{fig:5}       
\end{figure*}

\begin{table}
\caption{\leftline{Comparison with other models}}
\label{tab:4}       
\begin{tabular}{lll}
\hline\noalign{\smallskip}
Method & Arousal & Valence  \\
\noalign{\smallskip}\hline\noalign{\smallskip}
Zoubi et al. [24] & 88.54\% & 84.63\%  \\
Piho and Tjahjadi [25] & 89.84\% & 89.61\% \\
Our method & 94.42\% & 94.02\% \\
\noalign{\smallskip}\hline
\end{tabular}
\end{table}

\subsection{Four-class task}
The emotion recognition classification results for four-class task are displayed in Table 5. From Table 5, we can know that our model performs good on four-class task. It can be seen that when the feature set is the combined feature set of EEG, EOG and EMG, the highest classification accuracy is obtained by using the ensemble classifier, which is 90.74. It can be clearly seen from Figure 6 that when the same classification model is selected and different feature sets are combined, the classification results have little effect; but when the same feature set combination and different classification models are selected, the classification results are more obviously different. Therefore, we can conclude that when four-class task is performed on arousal and valence, the selection of feature sets has less influence on the results, and the selection of classification models has a greater impact on the results. And while the feature set combined by EEG, EOG and EMG, the ensemble classifier is adopted, the model is the most stable.
\\ It can be also seen from Table 5 and Fig. 6 that for four types of emotional states, decision trees, random forests, and ensemble classifier models have the best recognition ability on HAHV, while KNN performs best on LAHV. For the recognition of four emotional states, the combination of these three physiological signals and the ensemble classifier model obtain the best classification accuracy, namely, 88.81 on LALV, 91.58 on LAHV, 90.96 on HALV, 91.22 on HAHV, respectively.

\begin{table}
\caption{\leftline{Experimental results for four-class task}}
\label{tab:5}       
\begin{tabular}{llllll}
\hline\noalign{\smallskip}
first & second & third &aa &bb & cc \\
\noalign{\smallskip}\hline\noalign{\smallskip}
EEG+ENS & 90.41\% & 89.15\%	& 91.11\% &	89.70\% & 91.16\%\\
EEG+KNN 80.66\% & 82.15\%	& 84.71\% &	77.96\% & 79.22\%\\
EEG+CART & 87.26\%	& 86.81\% &	86.39\% &	86.77\% &	88.34\%\\
EEG+RF & 77.94\%	& 75.69\%	& 73.15\% &	79.00\% &	81.33\%\\
EEG+EOG+ENS & 90.38\% &	89.23\%	& 90.48\% & 89.82\%	& 91.33\%\\
EEG+EOG+KNN & 81.14\% & 81.69\% & 84.82\% &	79.63\% &79.66\%\\
EEG+EOG+CART & 87.47\% & 86.58\% & 86.51\% & 87.80\% & 88.31\%\\
EEG+EOG+RF & 78.21\% & 78.12\% & 74.97\% & 77.45\% & 80.63\%\\
EEG+EMG+ENS & 90.42\% & 89.73\% & 90.46\% & 90.12\% & 90.98\%\\
EEG+EMG+KNN & 80.83\% & 81.96\% & 84.43\% & 79.14\% & 79.19\%\\
EEG+EMG+CART & 87.29\% & 87.08\% & 86.42\%	& 87.73\% & 87.63\%\\
EEG+EMG+RF & 78.68\% & 77.15\%	& 76.72\% & 78.50\% &80.78\%\\
EEG+EOG+EMG+ENS &90.74\% & 88.81\% & 91.58\% & 90.96\% & 91.22\%\\
EEG+EOG+EMG+KNN &80.98\% &	81.65\%&85.07\% & 79.00\% &	79.52\%\\
EEG+EOG+EMG+CART &87.43\%	& 87.54\% &	86.02\% & 87.69\% & 88.04\%\\
EEG+EOG+EMG+RF & 80.23\% &	79.62\% & 77.43\% &	78.76\% & 83.15\%\\
\noalign{\smallskip}\hline
\end{tabular}
\end{table}

\begin{figure*}
  \includegraphics[width=1.0\textwidth]{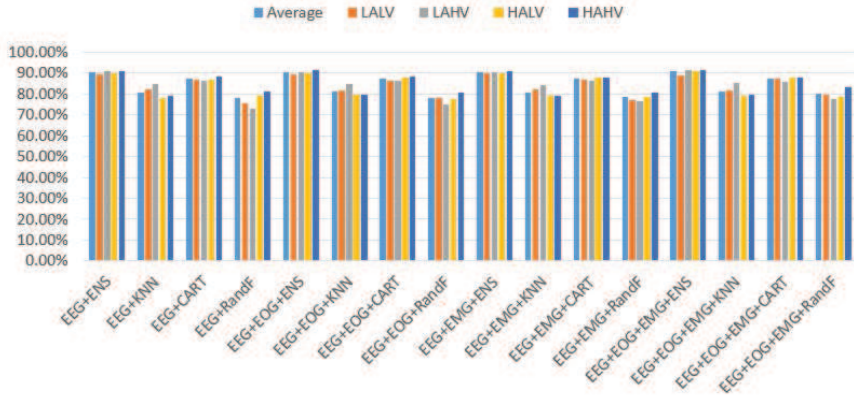}
\caption{Results of experiments for four-category task.}
\label{fig:6}       
\end{figure*}

\section{Conclusions}
In this paper, we propose an emotion recognition model mainly for two-class task and four-class task. By filtering signals firstly and computing parameters as the feature sets to classify, the experiment obtained accuracy of 94.42 and 94.20 for arousal and valence dimensions respectively for two-class task by the ensemble classifier with the combined of EEG, EOG, EMG. For four-class task, on the overall classification, 90.74 was obtained from the model. Both of them perform better than most existing methods. There are some other findings, comparing with the results of EEG data only, the addition of peripheral physiological signals gives better accuracy and makes the model more stable. Although, the different combination of physiological signals makes less effect than selecting different classifiers. The model we proposed has a good performance on classification emotional recognition task, but different people have different physiological characteristics, a model established for each person specially is supposed to be done. And other emotional database will be a part of on-going research to be added.

\section{Acknowledgment}
This work is partly supported by the National Natural Science Foundation of China (Nos. 61772252, 61976109, 61902165), the Natural Science Foundation of Liaoning Province of China (Nos. 2019-MS-216, 20180550542), the Doctoral Scientific Research Foundation of Liaoning Province (20170520207), Program for Liaoning Innovative Talents in University (No. LR2017044), and Dalian Science and Technology Innovation Fund (No.2018J12GX047).
\section{References}



\begin{thebibliography}{}
%
%
\bibitem{RefJA}
J. Deng, S. Fr¨¹hholz, Z. Zhang, and B. Schuller, ¡°Recognizing Emotions From Whispered Speech Based on Acoustic Feature Transfer Learning,¡± IEEE Access, vol. 5, pp. 5235-5246, Mar. 2017, doi: 10.1109/ACCESS.2017.2672722.
\bibitem{RefSB}
S. Shojaeilangari, W.Y. Yau, K. Nandakumar, J. Li, and E.K. Teoh, ¡°Robust Representation and Recognition of Facial Emotions Using Extreme Sparse Learning¡±, IEEE Transactions on Image Processing, vol. 24, no. 7, pp. 2140-2152, Mar. 2015, doi: 10.1109/TIP.2015.2416634.
\bibitem{RefZC}
Z. Yang and S.S. Narayanan, ¡°Modeling Dynamics of Expressive Body Gestures In Dyadic Interactions,¡± IEEE Transactions on Affective Computing, vol. 8, no. 3, pp. 369-381, 2017, doi: 10.1109/TAFFC.2016.2542812.
\bibitem{RefAD}
A.J. Casson, D.C. Yates, S.J. Smith, J.S. Duncan, and E. Rodriguez-Villegas, ¡°Wearable electroencephalography,¡± IEEE Eng Med Biol Mag, vol. 29, no. 3, pp. 44-56, 2010.
\bibitem{RefJE}
J.P. Rajan and S.E. Rajan, ¡°An internet of things based physiological signal monitoring and receiving system for virtual enhanced health care network,¡± Technology and Health Care, vol. 26, no. 2, pp. 379-385, Jan. 2018.
\bibitem{RefCF}
C. Maaoui and A. Pruski, ¡°Unsupervised stress detection from remote physiological signal¡±, Proc. IEEE International Conference on Industrial Technology (ICIT), Feb. 2018, doi: 10.1109/ICIT.2018.8352409.
\bibitem{RefJJ}
J. He and B. Han, ¡°Non-contact sleep staging algorithm based on physiological signal monitoring,¡± Proc. 4th World Conference on Control, Electronics and Computer Engineering (WCCECE 2018), pp. 308-314, Feb. 2018, doi: 10.25236/wccece.2018.63.
\bibitem{RefSH}
S.M. Alarcao and M.J. Fonseca, ¡°Emotions recognition using EEG signals: A Survey,¡± IEEE Transactions on Affective Computing, preprint, 12 Jun. 2017, doi: 10.1109/TAFFC.2017.2714671. (PrePrint)
\bibitem{RefYI}
Y. Zhong, M. Zhao, and Y. Wang, ¡°Recognition of emotions using multimodal physiological signals and an ensemble deep learning model,¡± Computer Methods and Programs in Biomedicine, vol. 140, pp. 93-110, Mar. 2017.
\bibitem{RefSG}
S. Jerritta, M. Murugappan, K. Wan, and S. Yaacob, ¡°Emotion recognition from facial EMG signals using higher order statistics and principal component analysis,¡± Journal of the Chinese Institute of Engineers, vol. 37, no. 3, pp. 385-394, 2014.
\bibitem{RefMK}
M. Nabian, Y. Yin, J. Wormwood, K.S. Quigley, L.F. Barrett, and S. Ostadabbas, ¡°An Open-Source Feature Extraction Tool for the Analysis of Peripheral Physiological Data,¡±. IEEE Journal of Translational Engineering in Health and Medicine, vol. 6, Oct. 2018, doi: 10.1109/JTEHM.2018.2878000.
\bibitem{RefSL}
S. Lin, J.Y. Xie, M.Y. Yang, and Z.Y. Li, ¡°A review of emotion recognition using physiological signals,¡± Sensors, vol. 18, no. 7, pp. 2074-2114, 2018.
\bibitem{RefYM}
Y. Wang, W. Zhang, L Wu, X Lin, X Zhao, Unsupervised metric fusion over multiview data by graph random walk-based cross-view diffusion, IEEE Transactions on Neural Networks and Learning Systems 28 (1), 57-70, 2017.
\bibitem{RefYN}
Y. Wang et al., Iterative Views Agreement: An Iterative Low-Rank based Structured Optimization Method to Multi-View Spectral Clustering, IJCAI 2016.
\bibitem{RefYO}
Y. Wang et al., Multiview Spectral Clustering via Structured Low-Rank Matrix Factorization. IEEE Trans. Neural Networks and Learning Systems, 2018.
\bibitem{RefWP}
W. Wen, G. Liu, N. Cheng, J. Wei, P. Shangguan, and W. Huang, ¡°Emotion recognition based on multi-variant correlation of physiological signals,¡± IEEE Transactions on Affective Computing, vol. 5, no. 2, pp. 126-140, 2014.
\bibitem{RefPQ}
P. Das, A. Khasnobish, and D.N. Tibarewala, ¡°Emotion recognition employing ECG and GSR signals as markers of ANS,¡± Proc. Conference on the Advances in Signal Processing, Pune, India, pp. 37-42, 2016.
\bibitem{RefWR}
W.Y. Hsu, ¡°EEG-based motor imagery classification using neuro-fuzzy prediction and wavelet fractal features,¡± Journal of Neuroscience Methods, vol. 189, no. 2, pp. 295-302, June 2010.
\bibitem{RefZS}
Z. Mohammadi, J. Frounchi, and M. Amiri, ¡°Wavelet-based emotion recognition system using EEG signal,¡± Neural Computing and Applications, vol. 28, no. 8, pp. 1985-1990, Aug. 2017.
\bibitem{RefYT}
Y. Zhang, S.H. Zhang, and X.M. Ji, ¡°EEG-based classification of emotions using empirical mode decomposition and autoregressive model,¡± Multimedia Tools and Applications, vol. 77, no. 20, pp. 26697-26710, Oct. 2018.
\bibitem{RefYU}
Y. Zhang, X.M. Ji, B. Liu, D. Huang, F.D. Xie, and Y.T. Zhang, ¡°Combined feature extraction method for classification of EEG signals,¡± Neural Computing and Applications, vol. 28, no. 11, pp. 3153-3161, 2017.
\bibitem{RefYV}
Y. Wang et al., Clustering via geometric median shift over Riemannian manifolds. Information Sciences 220, 292-305, 2013.
\bibitem{RefLW}
L. Wu, R. Hong, Y. Wang, M Wang. Cross-Entropy Adversarial View Adaptation for Person Re-identification. IEEE Transactions on Circuits and Systems for Video Technology, 2019.
\bibitem{RefLX}
L. Wu, Y. Wang, H. Yin et al., Few-Shot Deep Adversarial Learning for Video-based Person Re-identification, IEEE Transactions on Image Processing, 29 (1), 1233-1245, 2020.
\bibitem{RefLY}
L. Wu, Y. Wang, X. Li, J. Gao. Deep Attention-based Spatially Recursive Networks for Fine-Grained Visual Recognition. IEEE Transactions on Cybernetics 49 (5), 1791-1802, 2019.
\bibitem{RefLZ}
L. Wu, Y. Wang, L. Shao. Cycle-Consistent Deep Generative Hashing for Cross-Modal Retrieval. IEEE Transactions on Image Processing, 28 (4), 1602-1612, 2019.
\bibitem{RefZAA}
Z.H. Zhou, ¡°Ensemble Learning,¡± in: S.Z. Li, A.K. Jain (eds) Encyclopedia of Biometrics. Springer, Boston, MA, 2015.
\bibitem{RefSBB}
S. Koelstra, C. Muhl, and M. Soleymani, ¡°DEAP: A Database for Emotion Analysis Using Physiological Signals,¡± IEEE Transactions on Affective Computing, vol. 3, no. 1, pp. 18-31, Dec. 2011, doi: 10.1109/T-AFFC.2011.15.
\bibitem{RefNCC}
N.A. Pashtoon, ¡°IIR Digital Filters,¡± in: D.F. Elliott (eds) Handbook of Digital Signal Processing. Academic Press, pp. 289-357, 1987.
\bibitem{RefBDD}
B. Hjorth, "EEG analysis based on time domain properties," Electroencephalography and Clinical Neurophysiology, vol. 29, no. 3, pp. 306-310, Sep. 1970, doi: 10.1016/0013-4694(70)90143-4.
\bibitem{RefAEE}
A. Chatchinarat, K.W. Wong, and C.C. Fung, A comparison study on the relationship between the selection of EEG electrode channels and frequency bands used in classification for emotion recognition, Proc. International Conference on Machine Learning \& Cybernetics (ICMLC), pp. 251-256, February 2017, doi: 10.1109/ICMLC.2016.7860909.
\bibitem{RefOFF}
O.A. Zoubi, M. Awad, and N.K. Kasabov, ¡°Anytime multipurpose emotion recognition from EEG data using a Liquid State Machine based framework,¡± Artificial Intelligence in Medicine, vol. 86, pp. 1-8, Jan. 2018, doi: 10.1016/j.artmed.2018.01.001.
\bibitem{RefLGG}
L. Piho and T. Tjahjadi, ¡°A mutual information based adaptive windowing of informative EEG for emotion recognition,¡± IEEE Transactions on Affective Computing, May 2018, doi: 10.1109/TAFFC.2018.2840973. (PrePrint)
\end{thebibliography}
\end{document}